\documentclass[nohyperref]{article}

\usepackage{microtype}
\usepackage{graphicx}
\usepackage{subfigure}
\usepackage{booktabs} 

\usepackage{hyperref}
\usepackage{bm}
\usepackage[accepted]{icml2023}

\usepackage{amsmath}
\usepackage{amssymb}
\usepackage{mathtools}
\usepackage{amsthm}

\usepackage{multirow}
\usepackage{colortbl}
\usepackage{adjustbox}
\usepackage{setspace}   
\usepackage{enumitem}

\usepackage[capitalize,noabbrev]{cleveref}

\theoremstyle{plain}

\theoremstyle{definition}

\theoremstyle{remark}

\usepackage[textsize=tiny]{todonotes}
\definecolor{Gray}{gray}{0.9}

\icmltitlerunning{Chemically Transferable Generative Backmapping of Coarse-Grained Proteins}

\begin{document}

\twocolumn[
\icmltitle{Chemically Transferable Generative Backmapping of Coarse-Grained Proteins}

\icmlsetsymbol{equal}{*}

\begin{icmlauthorlist}
\icmlauthor{Soojung Yang}{1}
\icmlauthor{Rafael G\'{o}mez-Bombarelli}{2}
\end{icmlauthorlist}

\icmlaffiliation{1}{Computational and Systems Biology, MIT, Cambridge, MA, United States}
\icmlaffiliation{2}{Department of Material Science and Engineering, MIT, Cambridge, MA, United States}
\icmlcorrespondingauthor{Rafael G\'{o}mez-Bombarelli}{rafagb@mit.edu}
\icmlkeywords{Machine Learning, ICML, conformer generation, coarse-graining, protein structure prediction, equivariant neural network}

\vskip 0.3in
]

\printAffiliationsAndNotice{}

\begin{abstract}
Coarse-graining (CG) accelerates molecular simulations of protein dynamics by simulating sets of atoms as singular beads. Backmapping is the opposite operation of bringing lost atomistic details back from the CG representation. While machine learning (ML) has produced accurate and efficient CG simulations of proteins, fast and reliable backmapping remains a challenge. Rule-based methods produce poor all-atom geometries, needing computationally costly refinement through additional simulations. Recently proposed ML approaches outperform traditional baselines but are not transferable between proteins and sometimes generate unphysical atom placements with steric clashes and implausible torsion angles. This work addresses both issues to build a fast, transferable, and reliable generative backmapping tool for CG protein representations. We achieve generalization and reliability through a combined set of innovations: representation based on internal coordinates; an equivariant encoder/prior; a custom loss function that helps ensure local structure, global structure, and physical constraints; and expert curation of high-quality out-of-equilibrium protein data for training. Our results pave the way for out-of-the-box backmapping of coarse-grained simulations for arbitrary proteins. 
\end{abstract}

\section{Introduction}
\label{Introduction}
\begin{figure*}[ht] 
\vskip 0.2in
\begin{center}
\centerline{\includegraphics[width=1.8\columnwidth]{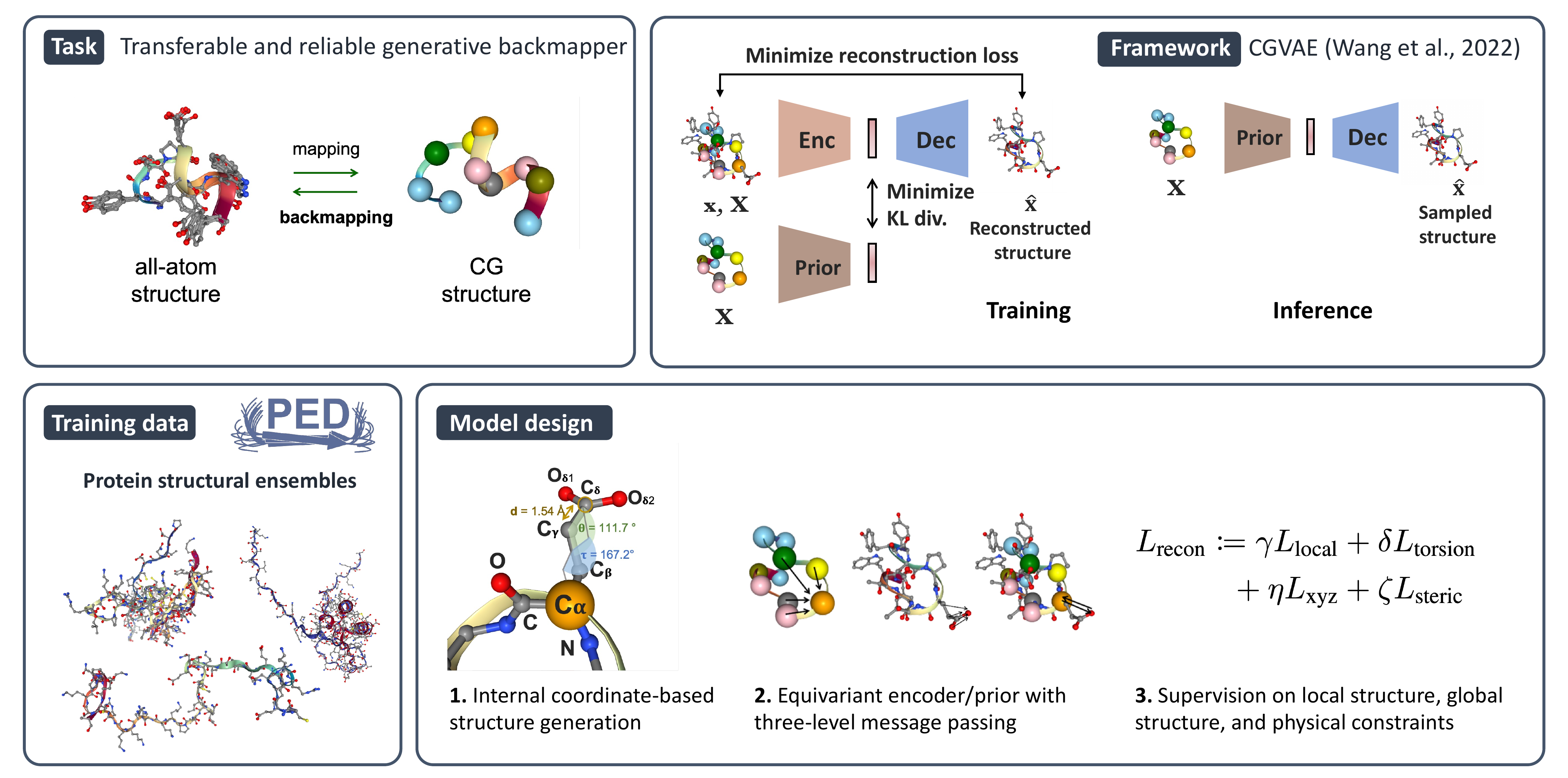}} 
\vspace{-0.02in}
\caption{Overview. We aim to build a transferable and reliable backmapping tool for proteins. Our method builds on a VAE framework \cite{Wang2022Generative}. We train the VAE model on the protein structural ensemble data curated from PED. Our model can be characterized with three components : internal coordinate-based representation, equivariant encoding, and physics-informed learning objectives. } 
\label{overview}
\end{center}
\vskip -0.2in
\vspace{-0.15in}
\end{figure*}

Protein dynamics ranges from large microsecond-scale movements of protein domains to small fast fluctuations of side chain atoms within protein pockets, and is connected to essential biological functions such as signaling, enzyme catalysis, and molecular machines \cite{Salvatella2014}. Despite the importance of dynamics and the large success of ML for prediction of protein structure, research on conformational ensembles started to accelerate only recently, mainly because data were scarce. Very flexible proteins (intrinsically disordered proteins, IDPs) or protein regions are better understood as conformational ensembles rather than static structures. Experimental structure determination methods observed either one frozen structure or the average of the conformational ensemble and thus are not suitable for describing individual dynamic states \cite{Miller2021}. Conformational ensembles are thus mainly generated using simulations such as Molecular Dynamics (MD) simulations or statistical sampling. Following the simulation, a representative subset can be selected from the pool of sampled conformers to match the properties and constraints derived from experimental measurements \cite{ConformerInsilico, Salvatella2014}.  

Atomistic simulations are often too computationally expensive for the time and length scale of protein dynamics. An effective way to overcome these limitations is to use coarse-grained (CG) simulations with simplified particles. Representing systems in a reduced number of degrees of freedom provides access to much larger spatiotemporal scales \cite{CG_chemreview}. However, the speedup comes at the cost of atomistic details, which are essential in determining protein biochemical functions. For example, identifying specific atom-level contacts at a protein-protein interaction surface or a ligand binding pocket is crucial to understand molecular recognition, signaling, or ligand binding \cite{BADACZEWSKADAWID2020162}. Thus, \textit{backmapping}, or restoring all-atom structures from CG structures, can be required to get a complete picture of protein function, especially for drug and protein design practices \cite{SLEDZ201893, Huang2016}.

Current popular backmapping methods involve two steps: 1) the generation of initial structures based on a set of geometric rules \cite{Lombardi2016-jy}, libraries of protein fragments \cite{Heath2007-gs}, or random placements \cite{randombackmap}, 2) the refinement of the generated structures by Monte Carlo relaxation or MD simulations. The second step is necessary because these rule-based sampling methods usually result in poor initial structures \cite{ROELTOURIS20201182}. However, the optimization step requires of an exhaustive computation and can be biased towards the choice of the scoring function and relaxation methods \cite{BADACZEWSKADAWID2020162}.  

Recently, data-driven methods have been proposed to achieve both efficiency and successful restoration of lost details through generative approaches. \citet{GANbackmap2020, Bereau2021, Wang2022Generative, TemporalBackmap2022} that learn the distribution of all-atom conformers conditioned on the CG structures. While those methods show promising performances on simple systems like alanine dipeptide and mini-protein chignolin, most methods cannot generalize beyond the chemistry on which they are trained \cite{GANbackmap2020, Wang2022Generative,TemporalBackmap2022}. \citet{Bereau2021} shows the possibility of chemical transferability by training the model on two small molecules and testing the model on a polymer whose monomers encompass each of the two small molecules. Still, no prior methods have been tested on structures that have high structural complexity and a wide range of flexibility as in large protein molecules.  

Here, we propose a deep generative backmapping tool that has transferability across protein space. Specifically, our model reconstructs the protein all-atom structure from the alpha carbon of each amino acid. We build the model on the framework of \citet{Wang2022Generative}, where a Variational Auto-Encoder (VAE) model approximates the 3D spatial distribution of all-atom structures conditioned on CG structures. We achieve the transferability by training on structures from the Protein Ensemble Database (PED) \cite{Lazar2021-ph}, which is a database of experimentally validated structural ensembles of IDPs and IDP-globular protein complexes. We hypothesize that a deep generative model, can learn the complex spatial interdependence of atoms and residues trained on a variety of geometries and chemical environments. We name our model \textbf{GenZProt}, as the model generates Z-matrix, a set of internal coordinates that defines a 3D molecular structure, for all-atom protein structures. GenZProt utilizes an equivariant encoder/prior that encodes residue-wise spatial information, and shows improved performance compared to its invariant counterpart and the ability to perform inference on arbitrary proteins outside the training dataset.

Naive rule- or ML-based backmapping strategies may fail to capture physical and chemical constraints, such as preserving the molecular connectivity of the all-atom representation, avoiding steric clashes, and reconstructing long-range interactions between side chains. GenZPort is constructed to preserve the topology by generating structures based on internal coordinates---bond length, bond angle, and torsion angle---instead of explicitly predicting Cartesian coordinates of atoms. Therefore, the training procedure relies on a loss function that optimizes local structure (bond length and bond angle), global structure (torsion angle and reconstruction in Cartesian space), as well as novel physical constraints (avoiding steric clashes). These design choices are proven to be crucial to achieve high-quality samples through ablation studies. 
We provide an overview of our method in Figure \ref{overview}.  

Our contributions can be summarized as follows:
\begin{itemize}[topsep=0.0pt,itemsep=1.2pt,leftmargin=5.5mm]
\item We propose the first data-driven generative backmapper that is transferable across the entire protein space. We achieve the transferability by training on computationally generated, experimentally validated diverse structural ensemble data.    
\item We propose a model design to achieve high-quality backmapping, relying on internal coordinates, an equivariant encoder, and loss functions that enforce physical constraints and preserve chemical connectivity. 
\end{itemize}

\section{Methods}

\subsection{Data}  
PED \cite{Lazar2021-ph} hosts 227 entries of protein structural ensembles, mostly computationally generated and experimentally constrained. Experimental validation reduces the potential bias introduce by errors in the sampling method, such as approximations in the force fields and thus provides better training statistics. From PED, we selected 84 proteins for training and four proteins for testing. The \textbf{Appendix} details the curation of training and testing set.  

\subsection{CG mapping scheme} We choose alpha Carbon ($C_{\alpha}$) mapping for coarse-graining--- every amino acid residue is represented as one bead centered at its $C_\alpha$. $C_{\alpha}$ atoms are explicitly present in popular medium resolution coarse-grained models, such as CABS (\citealp{Kolinski2004-nq}) or MARTINI (\citealp{MARTINI}). As a result, the majority of backmapping algorithms starts from the $C_\alpha$ trace level \citet{BADACZEWSKADAWID2020162}.  

\subsection{Internal Coordinate-based Structure Generation}
\vspace{-0.3in}
\begin{figure}[ht]
\vskip 0.2in
\begin{center}
\centerline{\includegraphics[width=\columnwidth]{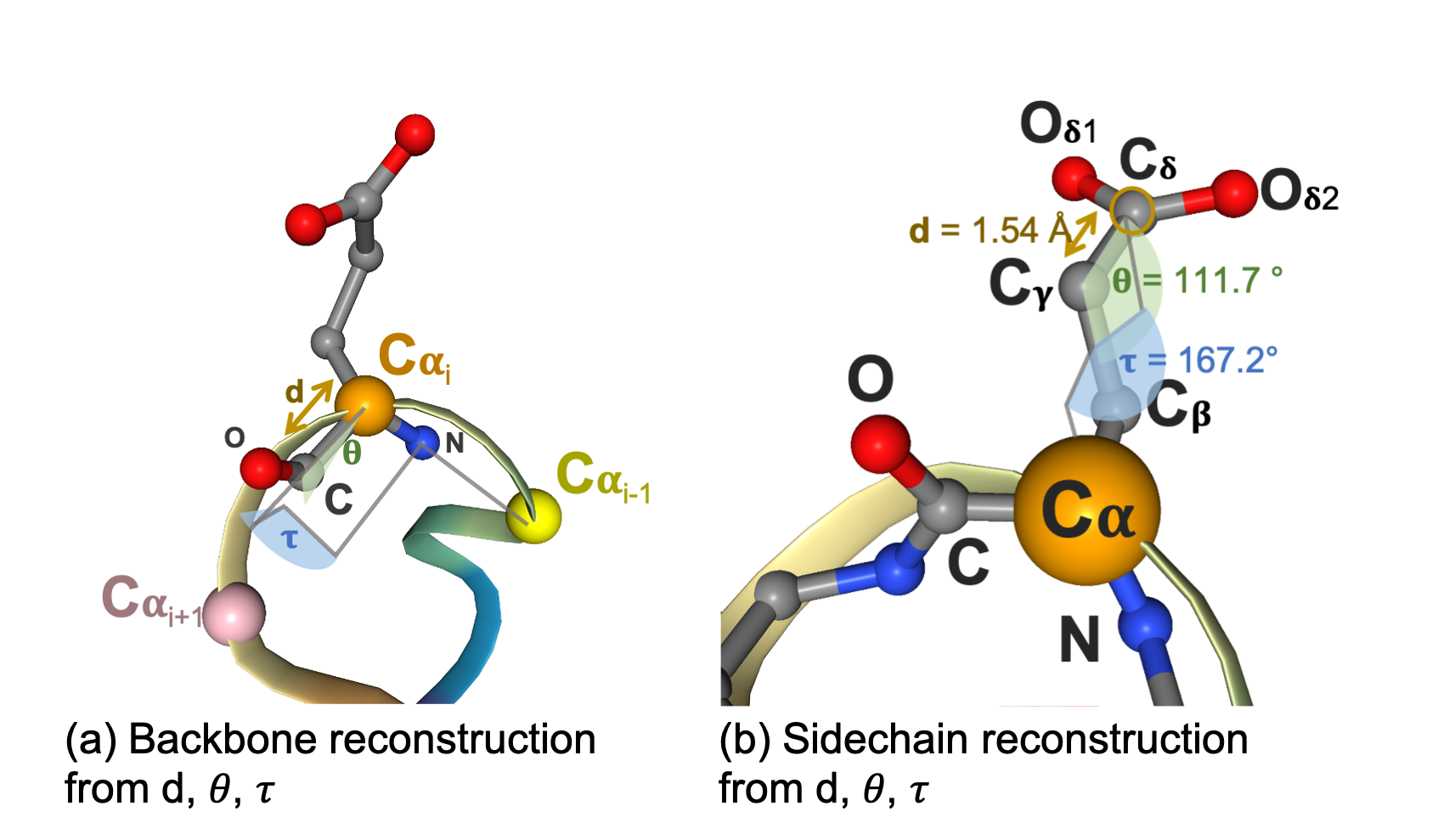}}
\caption{Internal coordinate-based reconstruction. \textbf{(a)} Backbone atoms $N_i, C_i$ are placed using adjacent three $C_\alpha$ as anchors. \textbf{(b)} Side chain atoms are placed using adjacent three atoms within the same residue. }
\label{ic_recon}
\end{center}
\vskip -0.2in
\end{figure}

Relying on internal coordinates makes it easier to preserve the bond topology, since bond lengths and angles, which are very sensitive to small distortions can be kept within a physical range. However, correctly predicting atomic placements and interactions in 3D space is as important as preserving the topology \cite{Lee2022}. Instead of attempting to reconstruct Cartesian coordinates, GenZProt achieves faithful reconstruction of the bond topology by generating internal coordinate representation of each atom directly as model outputs.  

GenZProt generates a set of internal coordinates (so-called Z-matrix), which is then converted to Cartesian coordinates through a rule-based algorithm. The placement of an atom $A$ in 3D space can be determined from three anchor atoms $B, C, D$ and a set of internal coordinates, bond length $d_{AB}$, bond angle $\theta_{ABC}$, and torsion angle $\tau_{ABCD}$, as shown in Figure \ref{ic_recon}.  

Since the topology of a residue is fully determined from its amino acid type, we use a predefined set of anchor atoms per-residue. However, the choice of the predefined set of anchors for $C_\alpha$ trace-to-all-atom backmapping task is not trivial. We devise a hierarchical atomic placement algorithm, where the backbone atoms are placed using $C_\alpha$s as anchors and the side chain atoms are placed sequentially.  

In \citet{Lombardi2016-jy} the authors postulate that the backbone atoms lie on the plane defined by three adjacent $C_\alpha$ atoms. Based on this assumption, we hypothesize that a machine learning model can learn to predict the placement of the backbone atoms of the $i^{th}$ residue, $N_i, C_i$, relative to three adjacent $C_\alpha$ atoms, $C_{\alpha_{i-1}}, C_{\alpha_i}, C_{\alpha_{i+1}}$. Once we obtain the placement of $C_{\alpha_i}, N_i, C_i$, we define three anchors within the $i^{th}$ residue to place a remaining backbone atom $O_i$ and side chain atoms. Atoms are then sequentially added to 3D space---for example, when the positions of $C_{\alpha_i}, N_i, C_i$ are known, $C_{\beta_i}$ is placed from the anchors $C_{\alpha_i}, N_i, C_i$, and with the $C_{\beta_i}$ position known, $C_{\gamma_i}$ is placed from anchors $C_{\beta_i}, C_{\alpha_i}, N_i$. We describe the transformation method in Figure \ref{ic_recon}.   

Despite the sequential transformation, our model has a short inference time since our decoder generates all internal coordinates simultaneously in one shot. Refer to \textbf{Appendix} for more details on Z-matrix to 3D coordinate conversion.   

\begin{figure*}[ht]
\vskip 0.2in
\begin{center}
\centerline{\includegraphics[width=1.5\columnwidth]{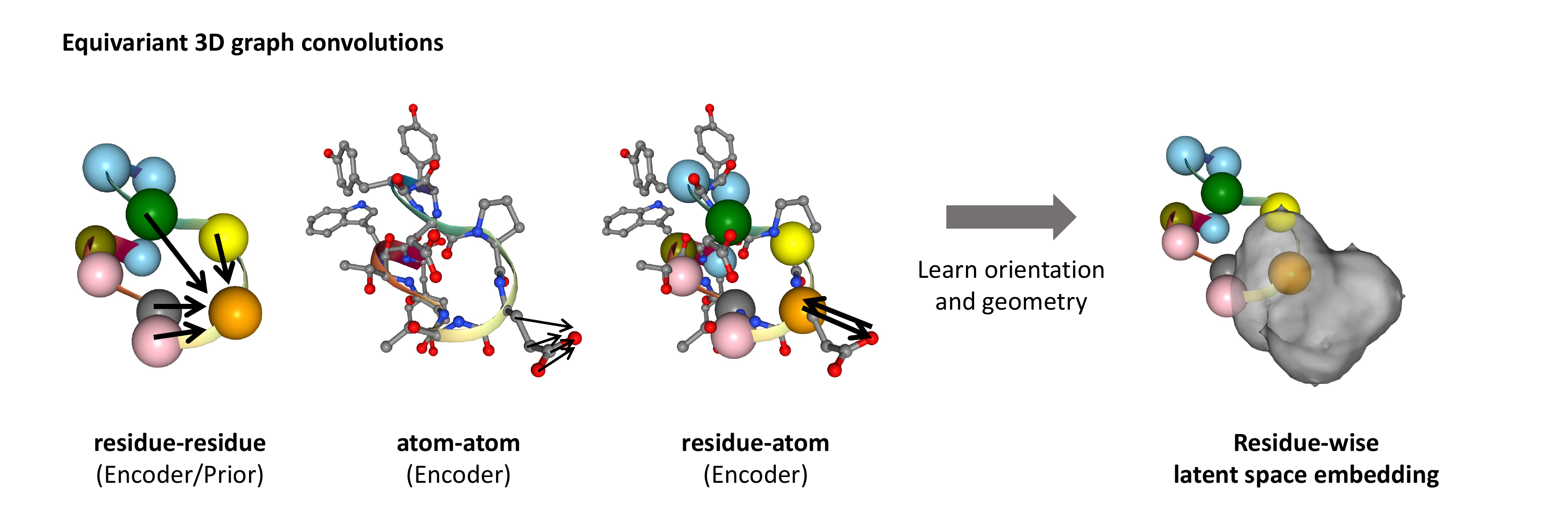}}
\caption{The three levels of equivariant 3D graph message passing operations in encoder and prior. }
\label{enc_prior}
\end{center}
\vskip -0.2in
\vspace{-0.1in}
\end{figure*}  

\subsection{VAE framework}
We build our model on the VAE framework introduced in \cite{Wang2022Generative}. In this framework, stochastic backmapping is formulated as a modeling task of the distribution of all-atom structure $x$ conditioned on CG structure $X$. 
The conditional distribution $p(x|X)$ is factorized as a latent variable model with a prior $P_\theta(z|X)$ and decoder $q_\psi(x|z, X)$, formulated as $p(x|X) \simeq q_\psi(x|z, X)P_\theta(z|X)$. 
The encoder $p_\phi(z|x, X)$ is introduced to train the learnable prior and decoder.     

During training, the CG latent variable $z$ is sampled from encoder $p_\phi(z|\mathbf{x}, \mathbf{X})$ as $z = \mu_\phi+\sigma_\phi \circ \bm{\epsilon}$, where $\bm{\epsilon} \sim N(0, I) $. During sampling, given the coarse structure $X$, we sample the latent variable from the prior ($z \sim P_\theta(z|\mathbf{X})$). Latent representation $z$ is then passed to the decoder to generate the all-atom structure $\hat{\mathbf{x}}$.  

\subsection{Model Architecture}
\textbf{Equivariant encoder and prior. }
We introduce an equivariant encoder and prior architecture designed to learn the spatial interdependence of atom and residue placements. Since it is intuitive to model molecular structures as graphs, we perform message passing operations on graphs where residues and atoms are the nodes. The orientation and geometry of the residues surrounding an atom are crucial to determine their 3D placement. Thus, we use geometric tensors to represent the node attributes and use SE(3)-equivariant neural networks to perform message passing on the nodes. This equivariant message passing neural network module was implemented with the \texttt{e3nn} library \cite{e3nn}, mainly referring to the score model of DiffDock \cite{corso2022diffdock}, which was used to predict docked poses of ligands in protein binding pockets.  

We digitize the protein molecular graph by assigning residue and atom identity as initial node attributes. In our model design, the encoder performs message passing at three levels: atom-atom pair within the cutoff distance $9 {\AA}$, atom-residue pair for every atom in a residue, and residue-residue pair within the cutoff distance $21 {\AA}$. The three levels of graph convolution are illustrated in Figure \ref{enc_prior}. The prior performs message passing at residue level only.   

\textbf{Invariant decoder. }
Molecular local structures---bond lengths and bond angles---are generally constrained to a single-mode Gaussian distribution with small variance, while torsion angles can vary more freely. Thus, we design a decoder architecture that allows flexibility on torsion angles and gives constrained predictions on local structures. Note that our backbone atom placement involves adjacent $C_\alpha$s, using angles $\theta_{NC_{\alpha_{i}}C_{\alpha_{i-1}}}$ and $\theta_{CC_{\alpha_{i}}C_{\alpha_{i+1}}}$. These angles have more variance than the side chain angles, so we also allow them more flexibility.  

We use a trainable lookup table to generate constrained variables such as bond lengths and side chain angles given a residue identity (PyTorch \texttt{nn.Embedding}). To predict the backbone angles and torsion angles, we perform message passing and pooling operations on node-wise feature vectors and then pass them to Multi-Layer Perceptron (MLP) layers.  

Indeed, a set of possible local structures of a molecule cannot be fully described by a lookup table. However, we found that the variance of angle distribution largely depends on the computational sampling method used to generate the ensemble. Thus, we removed the stochasticity in angles except for the ones involved in backbone placement.  

\begin{table*}[t]
\caption{\textbf{Ablation study on the model architecture}. \textbf{m1} : Our proposed model with equivariant encoder and invariant Z-matrix decoder. \textbf{m2} : Invariant encoder and Z-matrix decoder. \textbf{m3} : Equivariant encoder and Cartesian coordinate decoder. \textbf{m4} : Invariant encoder and Cartesian coordinate decoder. \textbf{m5} : \textbf{m1} trained with PED00151 only. \textbf{m6} : \textbf{m4} trained with PED00151 only. }
\label{tab:ab_arc}
\centering\small
\begin{tabular}{c l cccc}
        \toprule
        & Method & PED00055 & PED00090 & PED00151 & PED00218   \\
        \midrule
        \rowcolor{Gray}
         \multirow{6}{*}{RMSD ($\downarrow$)} 
         \cellcolor{white} & \textbf{m1 (GenZProt)}
         & \textbf{0.457}\scriptsize{$\pm$\textbf{0.002}}& \textbf{0.550}\scriptsize{$\pm$\textbf{0.005}} & \textbf{0.557}\scriptsize{$\pm$\textbf{0.001}} & \textbf{0.496}\scriptsize{$\pm$\textbf{0.001}}  \\
         & m2
         & 0.578\scriptsize{$\pm$0.004} & 0.787\scriptsize{$\pm$0.002} & 0.648\scriptsize{$\pm$0.005} & 0.565\scriptsize{$\pm$0.003}  \\
         & m3
         & 2.432\scriptsize{$\pm$0.035} & 2.475\scriptsize{$\pm$0.026} & 2.798\scriptsize{$\pm$0.011} & 2.393\scriptsize{$\pm$0.043}  \\
         & m4 (CGVAE) 
         & 2.244\scriptsize{$\pm$0.001} & 2.355\scriptsize{$\pm$0.002} & 2.901\scriptsize{$\pm$0.040} & 2.241\scriptsize{$\pm$0.004}  \\
         & m5 (GenZProt, single)
         & - & - & 0.832\scriptsize{$\pm$0.001} & - \\
         & m6 (CGVAE, single)
         & - & - & 2.072\scriptsize{$\pm$0.000} & - \\
         \midrule
          \rowcolor{Gray}
         \multirow{6}{*}{GED ($\downarrow$)} 
         \cellcolor{white} & \textbf{m1 (GenZProt)} 
         & \textbf{0.002}\scriptsize{$\pm$\textbf{0.000}} & \textbf{0.006}\scriptsize{$\pm$\textbf{0.000}} & \textbf{0.000}\scriptsize{$\pm$\textbf{0.000}} & \textbf{0.001}\scriptsize{$\pm$\textbf{0.000}}  \\
         & m2 
         & 0.007\scriptsize{$\pm$0.000} & 0.017\scriptsize{$\pm$0.000} & 0.005\scriptsize{$\pm$0.000} & 0.003\scriptsize{$\pm$0.000}  \\
         & m3 
         & 0.349\scriptsize{$\pm$0.035} & 0.431\scriptsize{$\pm$0.010} & 0.405\scriptsize{$\pm$0.002} & 0.339\scriptsize{$\pm$0.008}  \\
         & m4 (CGVAE)  
         & 0.246\scriptsize{$\pm$0.002} & 0.382\scriptsize{$\pm$0.004} & 0.308\scriptsize{$\pm$0.003} & 0.208\scriptsize{$\pm$0.002}  \\
        & m5 (GenZProt, single)
         & - & - & 0.084\scriptsize{$\pm$0.001} & - \\
         & m6 (CGVAE, single)
         & - & - & 0.140\scriptsize{$\pm$0.000} & - \\
         \midrule
        \rowcolor{Gray}
         \multirow{6}{*}{Steric clash ratio (\%; $\downarrow$)}          \cellcolor{white} & \textbf{m1 (GenZProt)}
         & \textbf{0.140}\scriptsize{$\pm$\textbf{0.003}} & \textbf{0.142}\scriptsize{$\pm$\textbf{0.002}} & \textbf{0.211}\scriptsize{$\pm$\textbf{0.008}} & \textbf{0.190}\scriptsize{$\pm$\textbf{0.003}}  \\
         & m2 
         & 0.173\scriptsize{$\pm$0.000} & 0.180\scriptsize{$\pm$0.003} & 0.267\scriptsize{$\pm$0.002} & 0.204\scriptsize{$\pm$0.002}  \\
         & m3 
         & 2.880\scriptsize{$\pm$0.622} & 3.517\scriptsize{$\pm$0.731} & 3.584\scriptsize{$\pm$0.362} & 3.088\scriptsize{$\pm$0.351}  \\
         & m4 (CGVAE)  
         & 1.880\scriptsize{$\pm$0.075} & 2.646\scriptsize{$\pm$0.046} & 3.027\scriptsize{$\pm$0.063} & 1.909\scriptsize{$\pm$0.012}  \\
        & m5 (GenZProt, single)
         & - & - & 1.090\scriptsize{$\pm$0.164} & - \\
         & m6 (CGVAE, single)
         & - & - & 2.032\scriptsize{$\pm$0.060} & - \\
        \bottomrule
\end{tabular}
\end{table*}

\begin{table*}[t]
\caption{\textbf{Ablation study on the reconstruction loss definition}. \textbf{m1} : Our proposed model with $L_\text{recon}$ defined in Equation \eqref{reconloss}. \textbf{m7} : Trained without $L_\text{torsion}$. \textbf{m8} : Trained without $L_\text{xyz}$. \textbf{m8} : Trained without $L_\text{steric}$ }
\label{tab:ab_loss}
\centering\small
\begin{tabular}{c l cccc}
        \toprule
        & $L_\text{recon}$ & PED00055 & PED00090 & PED00151 & PED00218   \\
        \midrule
         \rowcolor{Gray}
         \multirow{4}{*}{RMSD ($\downarrow$)} 
         \cellcolor{white} & \textbf{m1 (GenZProt)}
         & \textbf{0.457}\scriptsize{$\pm$\textbf{0.002}}& \textbf{0.550}\scriptsize{$\pm$\textbf{0.005}} & \textbf{0.557}\scriptsize{$\pm$\textbf{0.001}} & \textbf{0.496}\scriptsize{$\pm$\textbf{0.001}}  \\
         & m7 ($-L_\text{torsion}$)
         & 0.495\scriptsize{$\pm$0.002} & 0.582\scriptsize{$\pm$0.003} & 0.571\scriptsize{$\pm$0.001} & 0.509\scriptsize{$\pm$0.000}  \\
         & m8 ($-L_\text{xyz}$)
         & 1.910\scriptsize{$\pm$0.251} & 1.905\scriptsize{$\pm$0.136} & 2.025\scriptsize{$\pm$0.337} & 1.754\scriptsize{$\pm$0.198}  \\
         & m9 ($-L_\text{steric}$)
         & 0.467\scriptsize{$\pm$0.005} & 0.573\scriptsize{$\pm$0.013} & 0.570\scriptsize{$\pm$0.005} & 0.524\scriptsize{$\pm$0.003}  \\

         \midrule
          \rowcolor{Gray}
         \multirow{4}{*}{GED ($\downarrow$)} 
         \cellcolor{white} & \textbf{m1 (GenZProt)} 
         & 0.002\scriptsize{$\pm$0.000} & 0.006\scriptsize{$\pm$0.000} & \textbf{0.000}\scriptsize{$\pm$\textbf{0.000}} & \textbf{0.001}\scriptsize{$\pm$\textbf{0.000}}  \\
         & m7 ($-L_\text{torsion}$)
         & \textbf{0.001}\scriptsize{$\pm$\textbf{0.000}} & \textbf{0.004}\scriptsize{$\pm$\textbf{0.000}} & \textbf{0.000}\scriptsize{$\pm$\textbf{0.000}} & \textbf{0.001}\scriptsize{$\pm$\textbf{0.000}}  \\
         & m8 ($-L_\text{xyz}$)
         & 0.046\scriptsize{$\pm$0.000} & 0.057\scriptsize{$\pm$0.001} & 0.026\scriptsize{$\pm$0.000} & 0.033\scriptsize{$\pm$0.000}  \\
         & m9 ($-L_\text{steric}$)  
         & 0.002\scriptsize{$\pm$0.000} & 0.006\scriptsize{$\pm$0.000} & 0.003\scriptsize{$\pm$0.000} & \textbf{0.001}\scriptsize{$\pm$\textbf{0.000}}  \\
         \midrule
           \rowcolor{Gray}
         \multirow{4}{*}{Steric clash ratio (\%; $\downarrow$)} 
         \cellcolor{white} & \textbf{m1 (GenZProt)}
         & 0.140\scriptsize{$\pm$0.003} & 0.142\scriptsize{$\pm$0.002} & \textbf{0.211}\scriptsize{$\pm$\textbf{0.008}} & 0.190\scriptsize{$\pm$0.003}  \\
         & m7 ($-L_\text{torsion}$)
         & \textbf{0.135}\scriptsize{$\pm$\textbf{0.002}} & \textbf{0.131}\scriptsize{$\pm$\textbf{0.001}} & 0.236\scriptsize{$\pm$0.013} & 0.181\scriptsize{$\pm$0.003}  \\
         & m8 ($-L_\text{xyz}$)
         & 0.147\scriptsize{$\pm$0.005} & 0.221\scriptsize{$\pm$0.009} & 0.253\scriptsize{$\pm$0.041} & \textbf{0.144}\scriptsize{$\pm$\textbf{0.007}}  \\
         & m9 ($-L_\text{steric}$)  
         & 0.156\scriptsize{$\pm$0.001} & 0.157\scriptsize{$\pm$0.004} & 0.266\scriptsize{$\pm$0.002} & 0.199\scriptsize{$\pm$0.001}  \\
    
        \bottomrule
\end{tabular}
\end{table*}

\textbf{Loss functions. }
The VAE model is trained to minimize the Evidential Lower Bound (ELBO) objective, which includes the reconstruction term to train the encoder and decoder and the Kullback–Leibler (KL) divergence term to minimize the difference between the prior and the encoder \cite{VAEKingma}, namely $L_\text{ELBO} \coloneqq L_\text{recon} + \beta L_\text{KL}$. 

To learn geometry and interactions at the atomic level while ensuring the validity of the generated structures, we supervise the model on both topology and atom placements in 3D space. Topology reconstruction is measured by a Mean-Squared-Error (MSE) loss term on bond lengths ($L_{\text{bond}}$) and a periodic angular loss term for angles ($L_{\text{angle}}$). We define $L_{\text{local}}$ as a sum of $L_{\text{bond}}$ and $L_{\text{angle}}$ with $\epsilon=10^{-7}$:   
\begin{equation*}
\label{local_loss}
\begin{split}
  \underbrace{\frac{1}{|B|}\sum_{b \in B}(b - \hat{b})^2}_{L_{\text{bond}}} 
   + \underbrace{\frac{1}{|A|}\sum_{\theta \in A} \sqrt{2 (1-\cos(\theta - \hat{\theta})) + \epsilon}}_{L_{\text{angle}}},
  \end{split} 
\end{equation*}  
where $B$ is a set of all bonds, $b$ and $\hat{b}$ are ground truth and predicted bond length respectively. $A$ is a set of all angles, $\theta$ and $\hat{\theta}$ are ground truth and predicted angle in radian.   

Defining good reconstruction of atom placements in 3D space is not trivial for a backmapping task. A trivial solution for our internal coordinate-based generation setting would be a periodic angular loss term for torsion angles. However, a torsion angle can have a larger effect on the overall structure than other torsion angles. For example, a rotation near $C_\alpha$ would change the residue geometry more than a rotation at the end of the side chain. However, a simple regression would place an equal weight on every torsion angle. Thus, we additionally introduce a root-mean-squared distance (RMSD) loss term in Cartesian coordinate space:  
\begin{equation}
\begin{split}
  L_\text{torsion} & \coloneqq \frac{1}{|T|}\sum_{\tau \in T} \sqrt{2 \times (1-\cos(\tau - \hat{\tau})) + \epsilon} \\
  L_\text{xyz} & \coloneqq \frac{1}{|N|} \sum_{\mathbf{x} \in N}||\mathbf{x}-\hat{\mathbf{x}}||^2_2
\end{split}
\end{equation}
where $T$ is a set of all torsion angles, $\tau$ and $\hat{\tau}$ are ground truth and predicted torsion angle, respectively. $N$ is a set of all atoms, $\mathbf{x}$ and $\hat{\mathbf{x}}$ are ground truth and predicted Cartesian coordinates of an atom, respectively.  
   
To put further constraints on the chemical validity of the structures, we introduce steric clash loss, $L_{\text{steric}}$, as an auxiliary learning objective, defined as: 
\begin{equation}
  \begin{aligned}
  L_\text{steric} \coloneqq \sum_{\mathbf{x} \in \mathcal{N}} {\sum_{\mathbf{y} \in \mathcal{B}_{r} (\mathbf{x})} \max(2.0 - ||\mathbf{x}-\mathbf{y}||^2_2,  0.0)} 
  \end{aligned}
\end{equation}
where $\mathcal{B}_{r} (\mathbf{x})$ is a set of atoms within the cutoff distance $r = 5.0$ {\AA} with atom $\mathbf{x}$. Minimizing $L_\text{steric}$ keeps the distance between any two nonbonded atom pairs larger than 2.0 {\AA}.   

The reconstruction term then becomes:  
\begin{equation}
\label{reconloss}
  \begin{aligned}
  L_\text{recon} \coloneqq \gamma L_\text{local} + \delta L_\text{torsion} + \eta L_\text{xyz} + \zeta L_\text{steric}.
  \end{aligned}
\end{equation}
Hyperparameters $\gamma, \delta, \eta, \zeta$ are set to 1.0, 1.0, 1.0, 3.0, respectively. We explore different hyperparameter settings in our ablation study.

\begin{figure*}[ht]
\vskip 0.2in
\begin{center}
\centerline{\includegraphics[width=2.0\columnwidth]{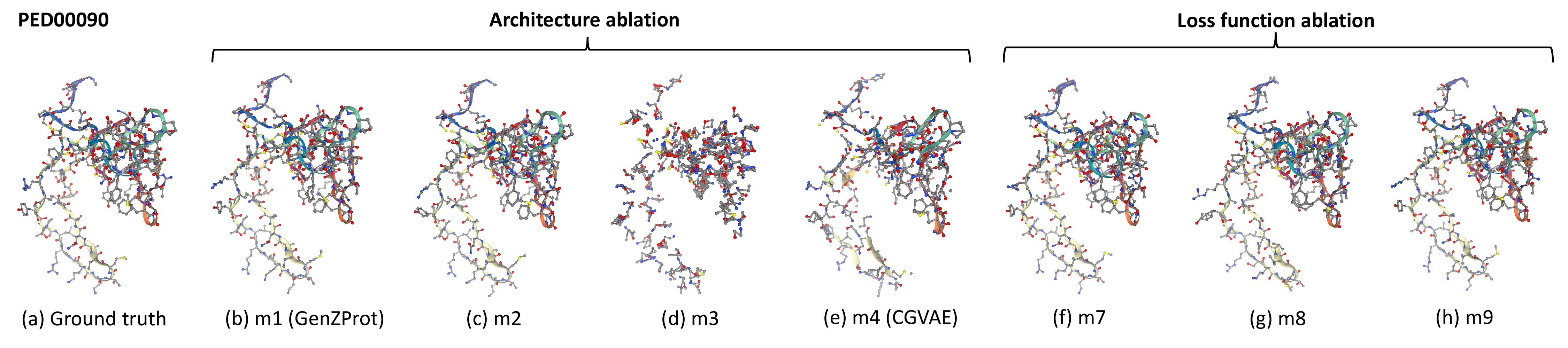}}
\caption{Reconstruction of PED00090 from \textbf{m1-m4}, \textbf{m7-m9}. }
\label{recon_90}
\end{center}
\vskip -0.3in
\end{figure*}

\begin{figure*}[h]
\vskip 0.2in
\begin{center}
\centerline{\includegraphics[width=1.5\columnwidth]{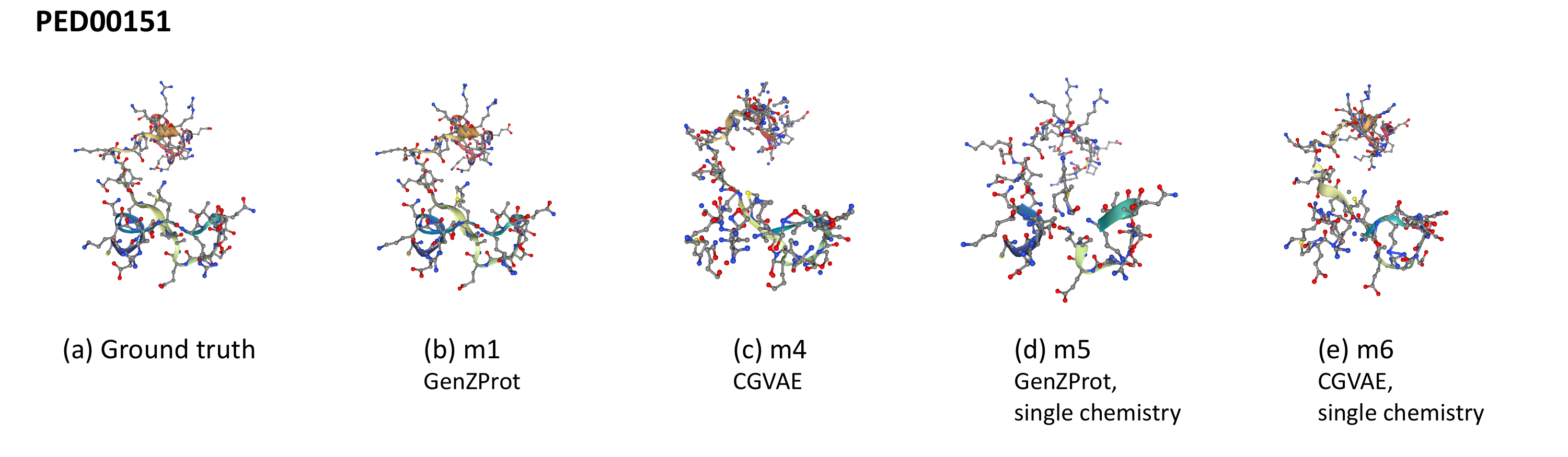}}
\caption{Reconstruction of PED00151 from transferable models \textbf{m1, m4} and single-chemistry models \textbf{m5, m6}.} 
\label{151transfer}
\end{center}
\vskip -0.2in
\end{figure*}

\section{Experiments}
In our experiments, we perform ablation studies on the model architecture and loss functions, and compare our model with the baseline, \textbf{CGVAE}. \textbf{CGVAE} was partially modified to take multiple proteins as training data. For each experiment, we perform five random seed experiments and report the mean and variance of the metrics. We refer the structures decoded from the encoder-sampled latent variables as \textit{reconstructed} and the structures generated from the prior sampling as \textit{sampled} structures.  

\subsection{Test Proteins}
We test our model with four proteins of varying flexibility and compactness: \textbf{PED00055} (87 residues), \textbf{PED00090} (92 residues), \textbf{PED00151} (46 residues), \textbf{PED00218} (129 residues). \textbf{PED00055} and \textbf{PED00090} are mostly globular with short disordered tails, \textbf{PED00151} is an IDP, and \textbf{PED00218} is a complex of a globular protein and an IDP.  

\subsection{Metrics}
We evaluate the model performance with three metrics: Root Mean Squared Distance (RMSD), Graph Edit Distance (GED), and Steric clash score.  

\textbf{Root Mean Squared Distance (RMSD).} To evaluate the reconstruction, we report the $RMSD$ value of ground truth and reconstructed structures for each model.  

\textbf{Graph Edit Distance (GED).} The sample quality is evaluated by measuring how well the generated geometries preserve the original chemical bond graph, which is quantified by the graph edit distance ratio $\lambda(G_{gen}, G_{true})$ between generated graph and the ground truth graph.  

\textbf{Steric clash score.} In addition to GED, we report the ratio of steric clash occurrence in all atom-atom pairs within a 5.0 {\AA} distance as a metric to measure the sample quality. For each atom-atom pair, distance smaller than 1.2 {\AA} is considered a steric clash.  

\section{Results}
\subsection{Ablation Studies}
\textbf{Transferability and model architecture. }

Table \ref{tab:ab_arc} shows how changing the model architecture affects the model performance (\textbf{m1-m6}). \textbf{m1-m4} are transferable models trained with 88 protein ensembles. \textbf{m5} and \textbf{m6} are single-chemistry models trained with \textbf{PED00151} alone.  

Our proposed model with equivariant encoder/prior and a Z-matrix decoder, \textbf{m1}, shows the best performance for every metric. \textbf{m1} performs better than the model with an invariant encoder/prior (\textbf{m2}), implying the importance of the encoder/prior equivariance. Models with a Cartesian coordinate decoder (\textbf{m3, m4}) fail to give high-quality reconstructions for our large test proteins. As shown in Figure \ref{recon_90}, reconstructions from \textbf{m3} and \textbf{m4} have many broken bonds and inaccurate topologies. Note that \textbf{m4} is equivalent to CGVAE, except that we modified its node definition to make it trainable for many proteins. We conclude that internal coordinate-based decoding coupled with equivariant encoder/prior can faithfully keep the topology while reconstructing high-quality structures with low RMSD and steric clash rates.  

We also analyze the effect of training on a large protein dataset compared to training on a single protein structure. \textbf{m5} has a model architecture identical to \textbf{m1} (GenZProt) while \textbf{m6} is identical to \textbf{m4} (CGVAE), except that \textbf{m5} and \textbf{m6} are trained with \textbf{PED00151} structures only (284 frames). \textbf{m1} performs better than \textbf{m5}, even though the training set does not include \textbf{PED00151}. Such a result proves that a generalized model could be a better choice for a structure with few data points than a single-chemistry model. \textbf{m6} performs better than its transferable version but still performs worse than internal coordinate-based models. Figure \ref{151transfer} is the visualization of the reconstructed structures from \textbf{m1}, \textbf{m4}, \textbf{m5}, and \textbf{m6}.  

\textbf{Learning objectives. }
In Table \ref{tab:ab_loss}, we evaluate the model performance as we change the learning objective. From \textbf{m7-m9}, the model architecture is identical to \textbf{m1}. Comparing \textbf{m1}, \textbf{m7}, and \textbf{m8}, $L_{xyz}$ was critical for optimal model performance and $L_{torsion}$ slightly improved the model performance. Furthermore, removing $L_{steric}$ resulted in increased steric clash ratio.     

\subsection{Qualitative analysis}

\begin{figure}[t!]
\vskip 0.2in
\begin{center}
\centerline{\includegraphics[width=0.55\columnwidth]{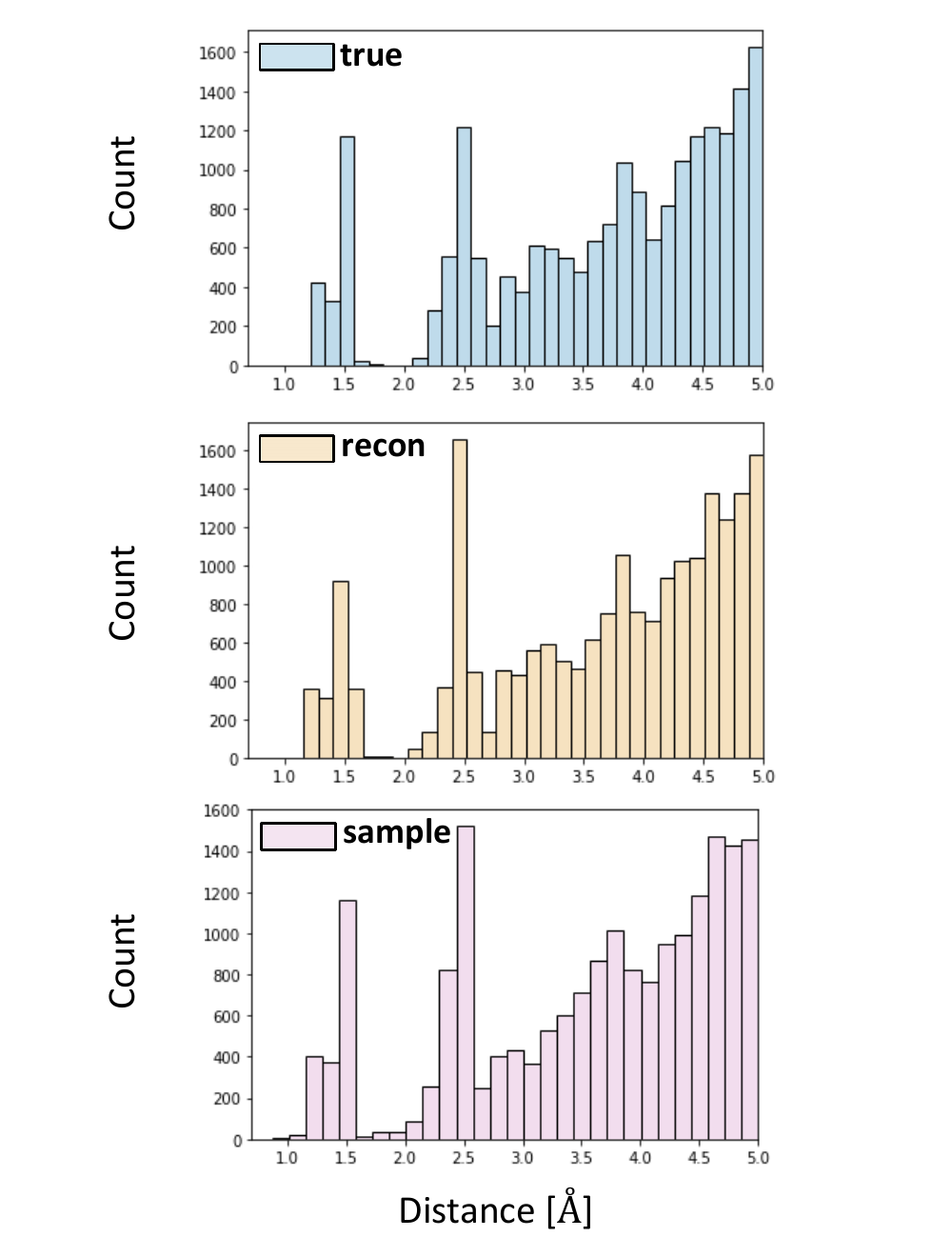}}
\caption{Atom-atom pairwise distances of ground truth, reconstructed, and sampled structures of PED00218.}
\label{distance_hist}
\end{center}
\vskip -0.2in
\vspace{-0.2in}
\end{figure} 

\textbf{Generated structures. }
Figure \ref{recon_all} shows reconstructed structures and sampled structures from \textbf{m1} (GenZProt) for four test proteins. Both reconstructed and sampled structures recover the topology faithfully and do not show any notable steric clashes.

\begin{figure}[t!]
\vskip 0.2in
\begin{center}
\centerline{\includegraphics[width=\columnwidth]{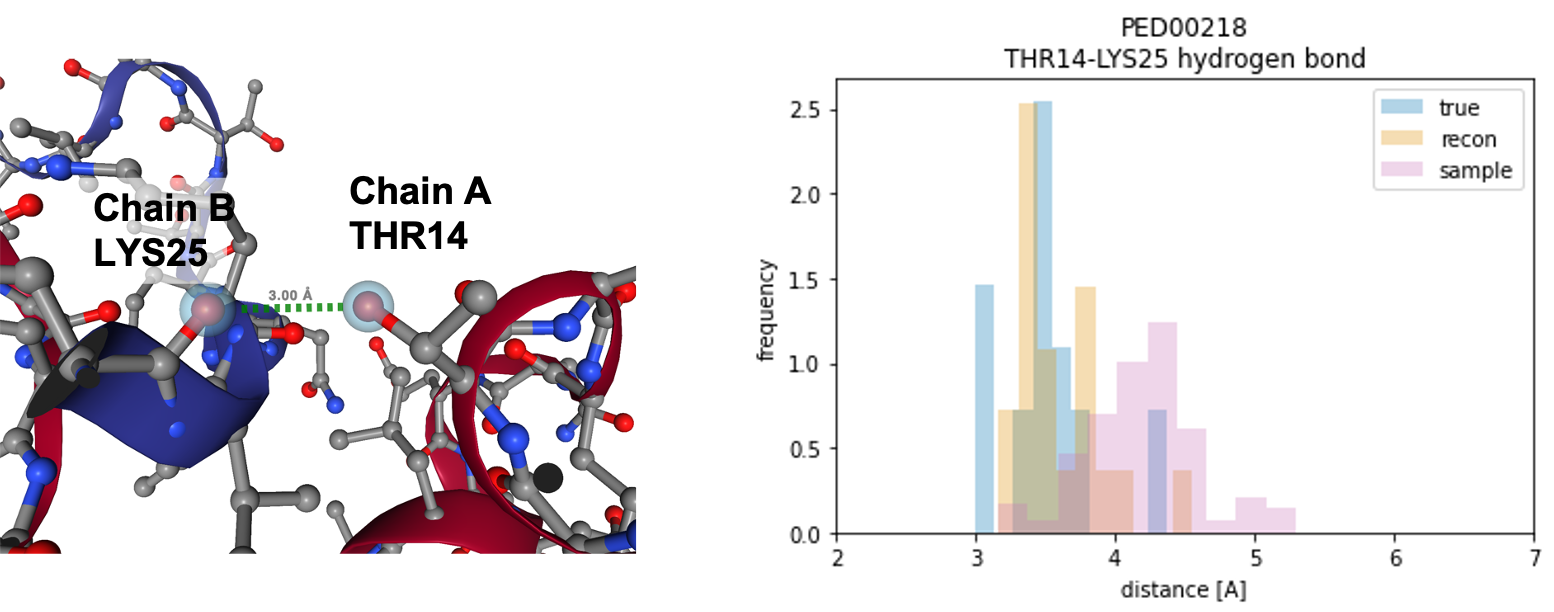}}
\caption{Histogram of distance between OG1 of THR14 (chain A) and atom O of LYS25 (chain B) in \textbf{PED00218}}
\label{inter_hist}
\end{center}
\vskip -0.3in
\end{figure} 

\begin{figure*}[]
\vskip 0.2in
\begin{center}
\centerline{\includegraphics[width=1.7\columnwidth]{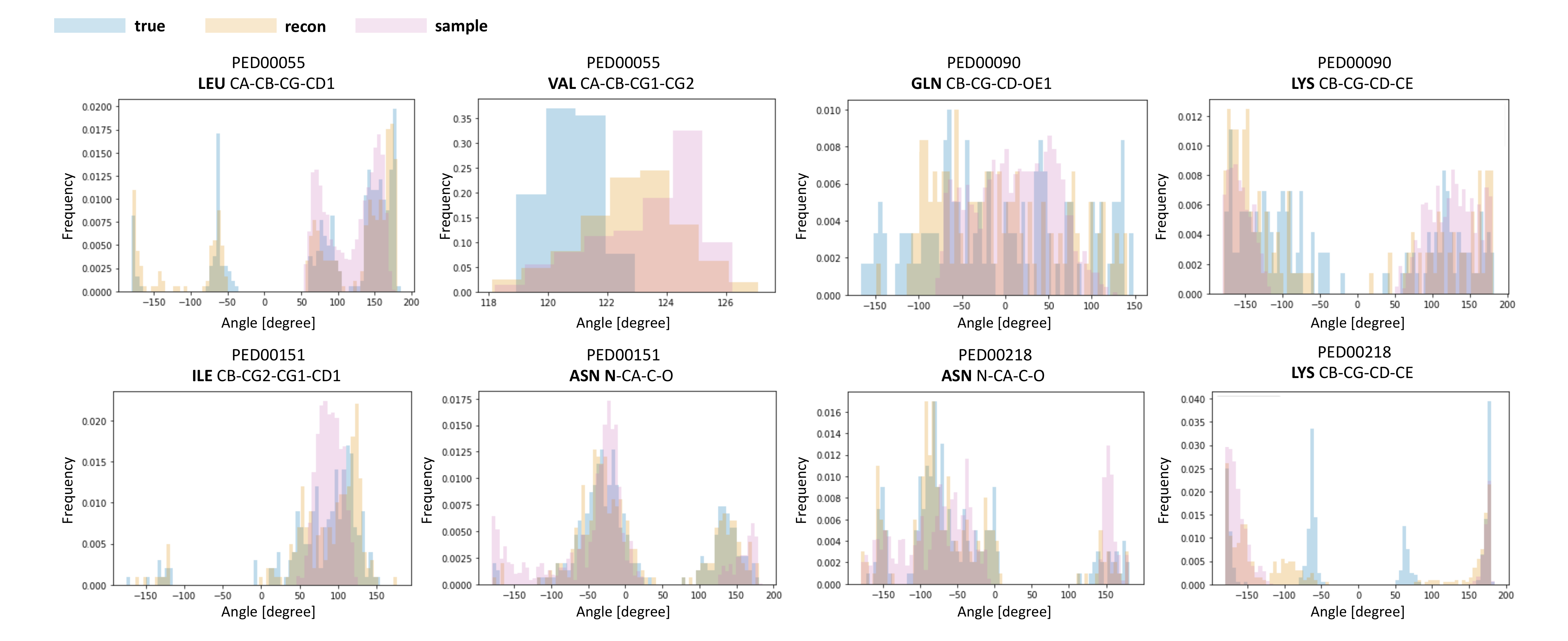}}
\caption{Histogram of torsion angles from the structures generated from \textbf{m1}. }
\label{torsion_hist}
\end{center}
\vskip -0.3in
\end{figure*}

\begin{figure*}[h]
\vskip 0.2in
\begin{center}
\centerline{\includegraphics[width=1.7\columnwidth]{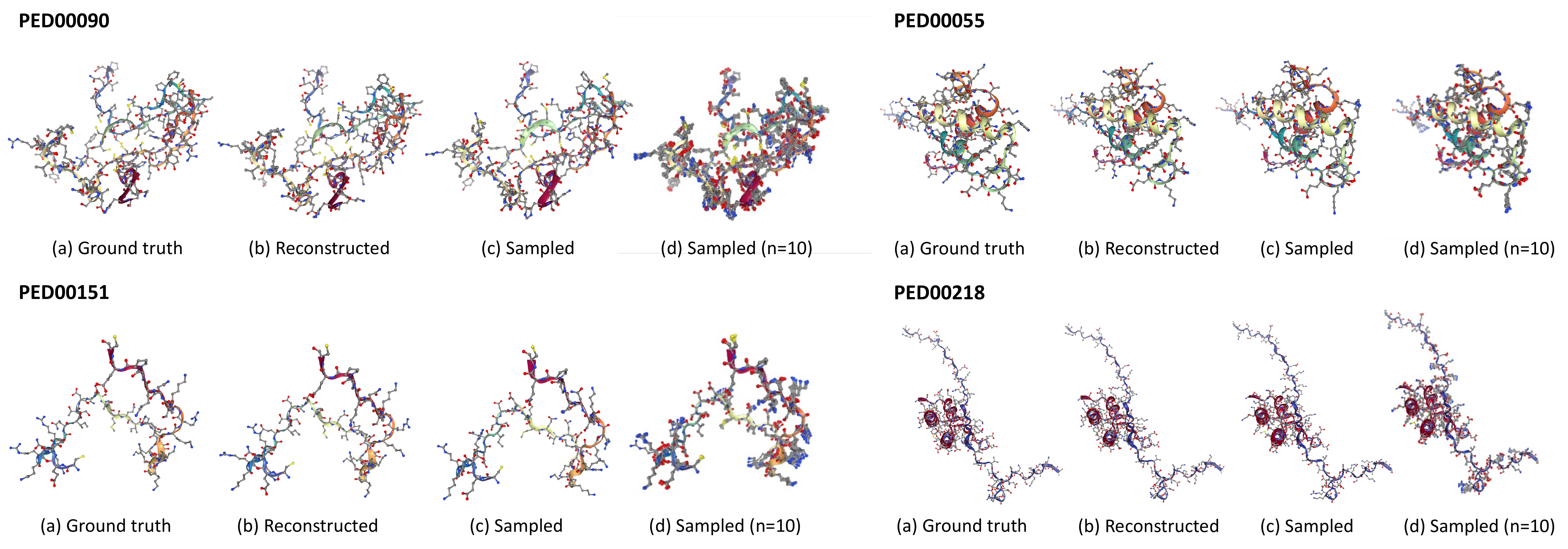}}
\caption{Structures reconstructed from \textbf{m1} for all four test proteins: PED00055, PED00090, PED00151, PED00218.}
\label{recon_all}
\end{center}
\vskip -0.3in
\end{figure*}

\textbf{Atom-atom distance distribution. } Figure \ref{distance_hist} shows all atom-atom pairwise distances $<5$ {\AA} in the ground truth structures (``true"), reconstructed structures (``recon"), and sampled structures (``sample") of \textbf{PED00218}, generated from \textbf{m1}. The ground truth distribution is well reconstructed by the encoder or the prior. 5 {\AA} is the higher cutoff for attractive London-van der Waals interactions \cite{Sengupta2012}. Encoder-generated reconstructions completely avoid steric clashes ($<1.2$ {\AA}), while prior-generated samples have few steric clashes. Atom-atom pairs with distance 3.3 {\AA} $< d <$ 4.0 {\AA} are likely hydrophobic interactions (van der Waals interactions), which is implied by a peak around 3.7 {\AA}. Both reconstructed and sampled structures have a peak at 3.7 {\AA} in a density similar to the ground truth, hinting that long-range interactions are preserved.  

We further investigate one long-range interaction in \textbf{PED00218}. \textbf{PED00218} is a peptide-protein complex, where a long IDP (chain B) is binding to a globular protein (chain A). The binding surface involves a hydrogen bond between a side chain oxygen of the chain A THR14 and a backbone oxygen of the chain B LYS25. Figure \ref{inter_hist} is the histogram of a distance between these two interacting atoms. The length of hydrogen bonds typically ranges in 2.7 {\AA} $< d <$ 3.3 {\AA} \cite{mcree_2012}. We can find reconstructed and sampled structures within the range of hydrogen bonds although the distributions are shifted to the right.  

\textbf{Torsion angle distribution. } Figure \ref{torsion_hist} shows the torsion angle distribution of ground truth, reconstructed, and sampled structures. Both the encoder- and prior-generated structures recover ground truth distributions well. However, as shown in the histogram for LYS of \textbf{PED00218}, the prior sometimes fails to find all modes of the distribution. This learning problem might be an inherent problem with VAE since its learning objective, a reverse KL divergence, can be minimized even when the prior fits to only one mode. As a result, the learned prior distribution would not spread out to low probability regions \cite{murphy_2012}. We propose to apply a diffusion model on the latent space of GenZProt as future work. Latent space diffusion, or stable diffusion, has been recently highlighted for achieving an expressive prior while retaining the generation quality \cite{diffusion}.

\section{Conclusion}
We introduce GenZProt, a transferable and reliable backmapper that can be used out-of-the-box for any arbitrary protein. We achieved chemical transferability by training on a protein conformational ensemble dataset curated from PED. Plus, we achieved reliability by employing physics-informed training objectives and devising an internal coordinate-based local structure construction method.  

As our model seamlessly handles an arbitrary number of peptide chains, our model can be utilized to repack side chains of protein binding interfaces. We showed the potential of using our model for binding surface reconstruction by testing on the protein-peptide complex PED00218. Upon binding or complex formation, protein side chain conformations can significantly change, and accounting for side chain flexibility can substantially improve protein-protein docking \cite{GRAY2003281}.  
   
Furthermore, in principle, our framework should be applicable to any family of polymers with a fixed number of building blocks. For future work, we propose applying our model to nucleic acids and nucleic acid-protein complexes.  

\section*{Software and Data}
Code and dataset for training and inference are available at \texttt{https://github.com/learningmatter-mit/GenZProt}. 

\section*{Acknowledgements}
We acknowledge the support from Novo Nordisk and Ilju Overseas Ph.D. scholarship. We thank Wujie Wang and Simon Axelrod for insightful discussion. We also thank Alexander Hoffman, Sihyun Yu, Akshay Subramanian, Yitong Tseo, and Lucia Vina Lopez for valuable feedback on the manuscript.  

\clearpage
\bibliography{reference}
\bibliographystyle{icml2023}

\newpage
\appendix
\label{appendix}
\onecolumn

\section{Speed Analysis} 
Our model shows fast sampling speeds of approximately 0.009 seconds per frame when tested with $\text{batch size} = 8$. The sampling time can be proportionally reduced as we increase the batch size.  

\begin{table*}[h]
\caption{Approximate inference times of GenZProt. }
\label{tab:app_time}
\centering\small
\begin{tabular}{c l c l c }
        \toprule
        protein & sequence length & time [sec/frame]  \\
        \midrule
        PED00055 & 87 &  0.006 \\
        PED00090 & 92 &  0.010 \\
        PED00151 & 46 & 0.006 \\
        PED00218 & 129 &  0.012 \\
        \bottomrule
\end{tabular}
\end{table*}  

\section{Related Work}
\label{Related Work}
Our work builds on \citet{Wang2022Generative}, among other recent studies on generative models for backmapping. \citet{Wang2022Generative} provides a principled probabilistic formulation of the backmapping problem and proposes CGVAE, a Variational Auto-Encoder (VAE) model that approximates the 3D spatial distribution of all-atom structures conditioned on CG structures. Compared to \citet{Wang2022Generative}, our work shows several significant advancements, including generalization to arbitrary proteins and faithful reconstruction of a protein's topology.  

Our work can be connected to protein structure prediction tasks. AlphaFold2 \cite{AlphaFold2021} showed that learning-based methods could give robust predictions for protein structures. However, AlphaFold2 is trained on crystallography-based structural data, which is mostly globular proteins. Also, the model is limited to a single structure prediction. To capture the ensemble of structures that characterizes a flexible biomolecule, one would need either new ML architectures trained on out-of-equilibrium data or MD simulations over large time scales. Our work explores both directions as we train our model on a database of IDP ensembles and test on backmapping tasks to assist CG MD simulation-based studies.    

While not specifically designed for backmapping, generative models have been used for small molecule conformer generation tasks. \citet{Jing2022Tor} connects to our work with its internal coordinate-based conformer generation framework, where bond lengths and angles are constrained, and torsion angles are predicted with a diffusion model. Note that we cannot directly use such models for protein backmapping tasks since a backmapper needs to be conditioned on CG structures. Also, small molecules have less complexity and fewer long-range interactions than macromolecules, meaning learning tasks for small molecules could be simpler.

\section{Background} 
Proteins are built from up to 20 different amino acids in Table \ref{aa}. In a protein chain, amino acids are connected to their neighbors by peptide bonds: an amide group of an amino acid forms a peptide bond (CO-NH) with a carboxyl group of an adjacent amino acid. Peptide bonds and the alpha carbons together form a continuous chain of atoms called \textit{backbone}. An individual amino acid connected to a peptide chain in a protein is called a \textit{residue}. Each residue has a chemical group attached to the alpha carbon, called \textit{side chain}.  

Proteins do not exist in a static snapshot but rather exist in ensembles of conformations. The common procedure of ensemble calculation involves the generation of a starting pool of conformations using sampling programs such as Flexible-Meccano (FM) \cite{FM}, TRaDES \cite{TRaDES}, or MD simulations. Then, a subset of conformers whose computed values fit the measurements from NMR or Small-Angle X-ray Scattering (SAXS) is selected as a representative structural ensemble. Each structure in a conformational ensemble is called a model or a frame.   

\begin{table*}[h]
\caption{Amino acid abbreviation chart}
\label{tab:aa}
\centering\small
\begin{tabular}{c c l c c}
        \toprule
Glycine	& G, GLY	&	Proline & P, PRO \\ 
Alanine	& A, ALA	&	Valine & V, VAL \\
Leucine	& L, LEU	&	Isoleucine & I, ILE \\
Methionine	& M, MET	&	Cysteine & C, CYS \\
Phenylalanine	& F, PHE	&	Tyrosine & Y, TYR \\
Typtophan	& W, TRP	&	Histidine & H, HIS \\
Lysine	& K, LYS	&	Arginine & R, ARG \\
Glutamine	& Q, GLN	&	Asparagine & N, ASN \\
Glutamic Acid	& E, GLU	&	Aspartic Acid & D, ASP \\
Serine	& S, SER	&	Threonine & T, THR \\
       \bottomrule
\end{tabular}
\end{table*}

\section{Training and Test Dataset}
Our training and test data are from the protein structural ensemble database PED \cite{Lazar2021-ph}. In this section, we discuss how we chose the entries for training and testing and provide analysis and statistics of the data.  

We split the train and test set by protein entries (i.e., models never see the test protein entries during training). The validation set is identical to the test set, and the learning rate reduction and early stopping are controlled based on the validation loss.  

\subsection{Training Proteins. } From 227 total entries of PED, we use 84 entries for training, four entries for validation, and four entries for testing.   

 The list of training entries are :   
 
PED00003, PED00004, PED00006, PED00011, PED00013, PED00022, PED00024, PED00025, PED00032, PED00033, PED00034, PED00036, PED00040, PED00041, PED00044, PED00045, PED00046, PED00050, PED00051, PED00052, PED00053, PED00054, PED00062, PED00072, PED00073, PED00074, PED00077, PED00078, PED00080, PED00085, PED00086, PED00087, PED00088, PED00092, PED00093, PED00094, PED00095, PED00097, PED00098, PED00099, PED00100, PED00101, PED00102, PED00104, PED00107, PED00109, PED00111, PED00112, PED00113, PED00114, PED00115, PED00117, PED00118, PED00120, PED00121, PED00123, PED00124, PED00125, PED00126, PED00132, PED00135, PED00141, PED00143, PED00145, PED00148, PED00150, PED00155, PED00156, PED00157, PED00158, PED00159, PED00160, PED00161, PED00175, PED00180, PED00181, PED00185, PED00190, PED00192, PED00193, PED00220, PED00217, PED00225, PED00227  

The list of validation entries are :  
PED00175, PED00023, PED00043, PED00119  

These proteins were excluded from the train and test set for the following reasons :  
\begin{itemize}
\item Metal ion-binding complexes : PED00009, PED00026, PED00035, PED00037, PED00038, PED00039, PED00058, PED00059, PED00063, PED00068, PED00069, PED00106, PED00108, PED00110, PED00131, PED00134, PED00136
\item Nucleotide-binding complexes : PED00057, PED00129, PED00130, PED00147
\item Cofactor-binding complexes : PED00075, PED00089, PED00091, PED00133, PED00222
\item PTM-including proteins except phosphorylation and oxidation : PED00014, PED00015, PED00047, PED00049, PED00064, PED00096, PED00127, PED00128
\item D-amino acid protein : PED00103
\item Proteins simulated or experimentally measured in unnatural conditions (e.g., denatured proteins, SDS or micelle containing solutions)  : PED00060, PED00061, PED00065, PED00066, PED00067, PED00081, PED00116, PED00144, PED00146, PED00147, PED00149, PED00152, PED00205
\end{itemize}
  
We included proteins with phosphorylation and oxidation PTM since they much more frequently appear than the other PTMs.   

Among 84 training entries, 23 entries were computed from the MD simulation. Sixty-one entries used sampling methods such as Flexible-Meccano, an all-atom structural optimization and sampling method for IDPs, based on amino acid-specific conformational potentials and volume exclusion \cite{FM}.   

\subsection{Test Proteins. }
In this section, we introduce our four test proteins : PED00055, PED00090, PED00151, and PED00218. Structural ensemble \textbf{PED00055}, the N-terminal domain of DNA polymerase $\beta$, is sampled with an X-PLOR \textit{ab initio} simulation and constrained with CHARMM parameters and NMR measurements. \textbf{PED00090} is a structural ensemble of the human chorionic gonadotropin alpha subunit sampled with X-PLOR and constrained with NMR measurements. \textbf{PED00151} is a structural ensemble of a Nuclear Localization Signal (NLS 99-140) peptide, sampled with MD simulation package CAMPARI and reweighted to match the experimental measurement from smFRET and SAXS. \textbf{PED00218} is a structure ensemble of a complex Taf14ET-Sth1EBMC, and its structures were derived from MD simulation and fit to NMR measurements. PED provides 55, 27, 29,598, 20 frames for entries PED00055, PED00090, PED00151, and PED00218, respectively. We use all frames for PED00055, PED00090, and PED00218 as testing set. For PED00151, we randomly sample 140 frames from the ensemble PED00151e000.   

\subsection{Single Chemistry Experiments}
We perform the single chemistry experiments with entry PED00151. PED provides three ensembles for PED00151 : PED00151e000 (9,746 frames), PED00151e001 (9,924 frames), and PED00151e002 (9,928 frames). Each ensemble is reweighted with the COPER program \cite{COPER} to match the experimental FRET efficiency and $R_g$ values. To reduce the training time, we randomly sample 140, 142, and 142 samples from the ensemble PED00151e000, PED00151e001, and PED00151e002, respectively. We use PED00151e001 and PED00151e002 samples (284 frames) as the train and validation set. We randomly select 224 frames as training set and use the remaining 60 frames for validation. PED00151e000 (140 frames) is used as the test set.  

\subsection{Data Statistics}
\begin{figure}[h] 
\vskip 0.2in
\begin{center}
\centerline{\includegraphics[width=0.8\columnwidth]{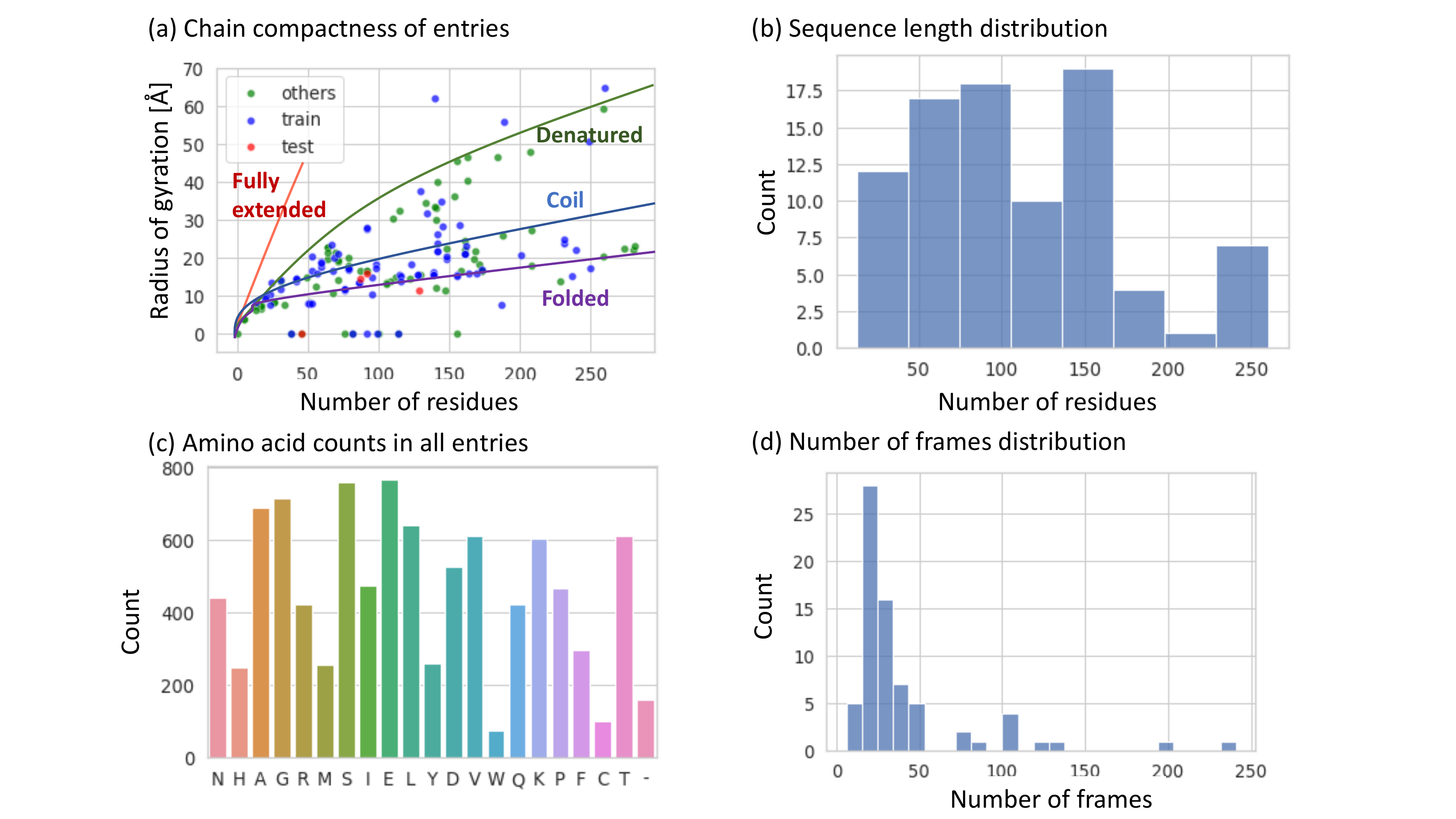}} 
\caption{\textbf{(a)} Compactness plot. Train set, test set, and excluded entries are colored in blue, red, and green, respectively. Large proteins (number of residues $>$ 300) are omitted.} 
\label{data_stat}
\end{center}
\vskip -0.2in
\end{figure}

This section provides a quantitative analysis of the train and the test set. Our training set includes $\sim$ 10,000 frames, and the test set includes $\sim$ 500 frames. Our training and test proteins have 9,562 and 354 residues in total, respectively. In other words, the model has seen $\sim$ 10,000 different residue environments. The distribution of protein sequence length and the number of frames are shown in Figure \ref{data_stat} plot \textbf{(b)} and \textbf{(d)}. Figure \ref{data_stat} plot \textbf{(c)} shows the distribution of amino acid counts in all training entries. The amino acids are well distributed, except for tryptophans (TRP; W) and cysteines (CYS; C).         

Our dataset includes proteins of various levels of compactness. Protein compactness can be characterized by the radius of gyration ($R_g$) as a function of the chain length \cite{Lazar2021-ph}. Figure \ref{data_stat} plot \textbf{(a)} plots $R_g$ of protein chains against the chain length. Each dot represents a chain in an entry. The trend lines in Figure \ref{data_stat} plot \textbf{(a)} are taken from Figure 2 of \citet{Lazar2021-ph}: completely flexible, rod-like chains follow a linear trend since the size of the protein will be proportional to the sequence length, folded proteins approximately follow a known scaling law, and disordered proteins fall in between. As shown in the plot, our test set does not include proteins with extreme disorderedness. However, since one of our test proteins, PED00151, is a disordered IDP with partial coils, we assume that testing on PED00151 would be enough to show the model performance on flexible proteins.      

\subsection{Data Preprocessing}
\textbf{Hydrogen removal. } Since residues can be in many protonation states, we remove all the hydrogens from the train and test structures to reduce the number of building block representations. Moreover, in practice, protonation and hydrogen placement software such as REDUCE (\citealp{REDUCE}) have been reliably used. Thus, we only consider heavy atoms for our reconstruction and sampling tasks.  

\textbf{Handling terminal residues and multiple chains. }
Since we reconstruct backbone nitrogens and carbons with three alpha carbons ($C_{\alpha_{i-1}}, C_{\alpha_{i}}, C_{\alpha_{i+1}}$ where $N > i \geq 1$) as anchors, we cannot reconstruct atomistic positions for terminal residues. Therefore, we mask the $i=0$ and $i=N$ residues for training and inference. Also, when the entry is a protein complex with multiple chains, two terminal residues exist for each chain. In such cases, we mask all the terminal residues.  

\textbf{Handling PTMs. } We treat phosphorylated Threonine (TPO) and phosphorylated Serine (SPO) as individual building blocks in addition to 20 canonical amino acids. We include proteins with oxidated residues (OXT) in our training and test sets. However, we do not treat oxidized residues as separate building blocks since oxidation appears in many amino acid types. Instead, we remove all the additional oxygen atoms added by oxidation PTM.  

\textbf{Sampling a subset from large entries. } For the entries with a large number of frames ($\text{Number of frames} > 500$), we use the sampled subset of the entry to avoid the model overrepresenting those entries. We sample so that the number of frames per entry does not exceed 500. Following is the list of the large entries : PED00003, PED00006, PED00011, PED00022, PED00024, PED00025, PED00143, PED00145, PED00148, PED00150, PED00155, PED00180, PED00181.  

\section{Molecular Geometry and Internal Coordinate System}
One possible representation of the molecular geometry is to list Cartesian coordinates of each atom. However, bond length, bond angle, and torsion angle are a more natural representation of proteins than the Cartesian coordinates since a topology of a molecule does not change unless it goes through a chemical reaction. In addition, since bond length, bond angle, and torsion angle have different frequencies of degrees of freedom, it could be easier to manipulate geometry and perform a conformational search with internal coordinate representation \cite{ICMD}.  

To fully specify molecular geometry with Cartesian coordinates, $3N$ values are needed for a system of $N$ atoms (i.e., $x, y, z$ for each atom). For internal coordinate-based representation, it is a convention to specify a molecular geometry with Z-matrix. Each line of the Z-matrix defines a position of an atom: $i, \text{atom type}, j, d_{ij}, k, \theta_{ijk}, l, \tau_{ijkl}$, where $i$ is the index of the current atom whose position is being defined, and $j, k, l$ are the indices of adjacent atoms whose positions are already defined. The positions of the atoms $j, k, l$ are used as anchors to place the atom $i$. $d$, $\theta$, and $\tau$ are distance, angle, and torsion angle, respectively. Thus, our decoder outputs three values per atom $i$, $d_{ij}, \theta_{ijk}, \tau_{ijkl}$, where indices $j, k, l$ are predefined given the residue type. During training, a fully differentiable Algorithm \ref{alg:atom_placement} is used to convert the Z-matrix to Cartesian coordinates. Then, $L_{xyz}$ and $L_{steric}$ are computed from the reconstructed Cartesian coordinate.  

\begin{figure}[h] 
\vskip 0.2in
\begin{center}
\centerline{\includegraphics[width=0.3\columnwidth]{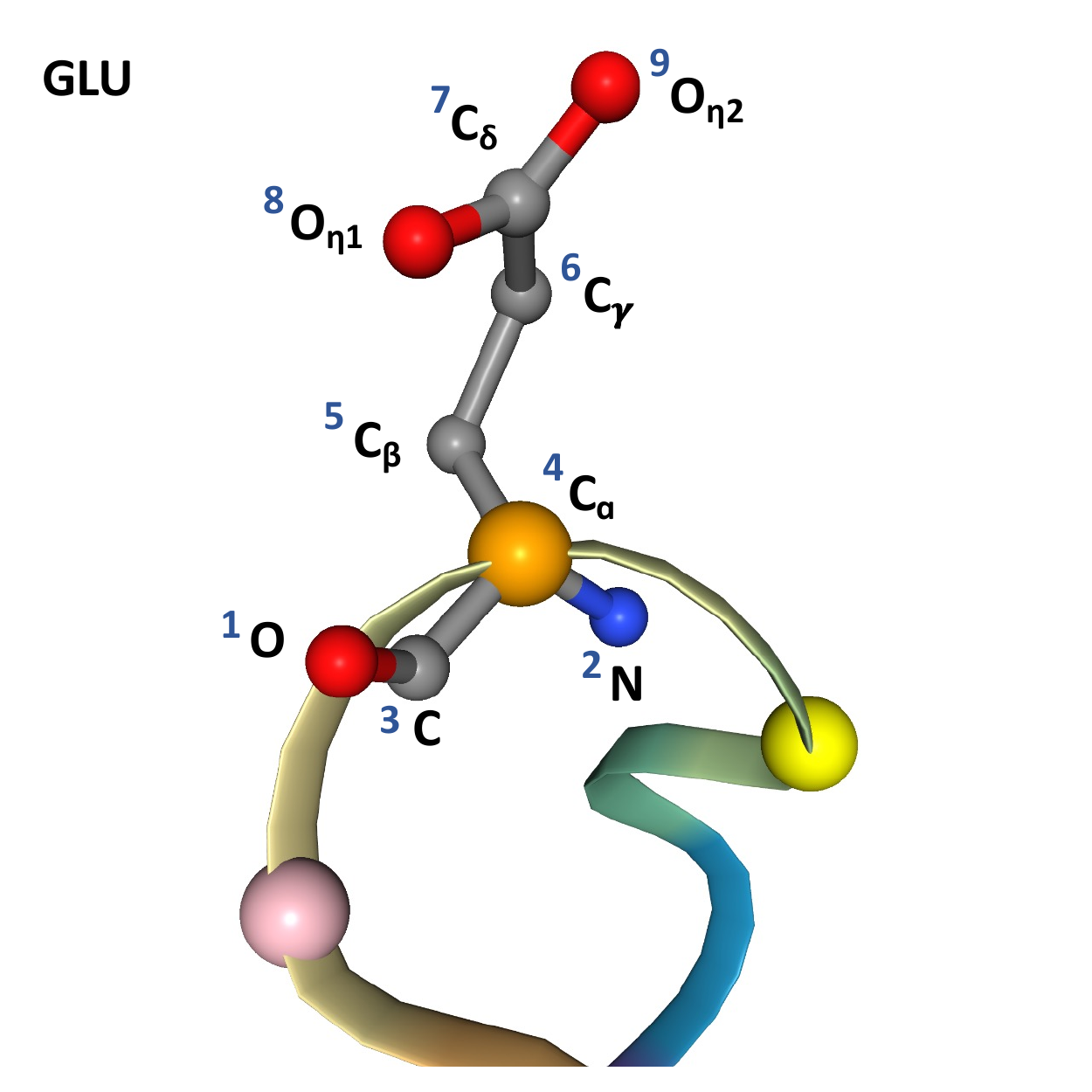}} 
\caption{Structure of a glutamic acid. } 
\label{ic_glu}
\end{center}
\vskip -0.2in
\end{figure}

Atoms in a residue are placed sequentially. For example, as shown in Figure \ref{ic_glu}, the beta carbon ($C_\beta$), $i=5$, is constructed from atoms $j=4, k=3, l=2$, which are the alpha carbon, $C$, and $N$, respectively. Similarly, the gamma carbon ($C_\gamma$, $i=6$) is constructed from atoms $j=5, k=4, l=3$, which correspond to the beta carbon, alpha carbon, and $C$, respectively.  

However, adding atoms one by one will require $N$ steps for a protein with $N$ atoms, which will be extremely time-consuming. Thus, we reconstruct all residues at once in a parallel manner. For the $i$th step of the conversion, $i$th atoms of all residue are placed simultaneously. The order of the atoms is predefined (e.g., $L=$[O, N, C, CA, CB, CG, CD, OE1, OE2] for GLU). For any protein, 13 conversion steps are executed, as the maximum number of atoms in a residue is 13 except the already known $C_\alpha$. 

\begin{algorithm}[tb]
   \caption{A pseudocode for the reconstruction of the list of Cartesian coordinates of side chain atoms, $L$, \\ for a residue with $m$ side chain atoms.}
   \label{alg:atom_placement}
\begin{algorithmic}
\STATE {\bfseries Input:} $L = [\mathbf{x}_1, \mathbf{x}_2, \mathbf{x}_3, \mathbf{x}_4]$,   {\small\color{gray}\tt \# $\mathbf{x}_1$, $\mathbf{x}_2$, $\mathbf{x}_3$, $\mathbf{x}_4$, correspond to $O, C, N, CA$, respectively}
\FOR{$i=5$ {\bfseries to} $m+4$}
   \STATE {\bfseries Input:} row $i$ of the Z-matrix $d_{ij}, \theta_{ijk}, \tau_{ijkl}$
   \STATE Let $j=i-1, k=i-2, l=i-3$
   \STATE Compute $\mathbf{v}_{jk} \coloneqq$ $L$[$j$] $- L$[$k$]
   \STATE Compute $\mathbf{v}_{kl} \coloneqq$ $L$[$k$] $- L$[$l$]
   \STATE Compute $\mathbf{v} \coloneqq d_{ij}\mathbf{v}_{jk} / \|\mathbf{v}_{jk}\|^2_2 $ 
    {\small\color{gray} \tt \# a vector of length $d_{ij}$ pointing from $j$ to $k$}
   \STATE Compute $\mathbf{n} \coloneqq \mathbf{v}_{jk} \times \mathbf{v}_{kl}$ {\small\color{gray} \tt \# a vector normal to the plane defined by $j, k, l$}
   \STATE $\mathbf{v} \leftarrow \mathbf{R}(\theta_{ijk})\mathbf{v}$ {\small\color{gray} \tt \# Rotate $\mathbf{v}$ around $n$ by $\theta_{ijk}$ }
   \STATE $\mathbf{v} \leftarrow \mathbf{R}(\tau_{ijkl})\mathbf{v}$ {\small\color{gray} \tt \# Rotate $\mathbf{v}$ around $\mathbf{v}_{jk}$ by $\tau_{ijkl}$}
   \STATE $L$[$i$] = $\mathbf{v}$ $+ L$[$j$] {\small\color{gray} \tt \# Cartesian coordinate of $i$th atom}
\ENDFOR
    \STATE {\bfseries Return} $L$
\end{algorithmic}
\end{algorithm}

\section{Experimental Details}
\subsection{GenZProt}
Our proposed model and the ablation models are trained with the hyperparameters defined in Table \ref{tab:app_hyperparam}. Models were trained with Xeon-G6 GPU nodes until convergence, with a maximum runtime of 20 hours. Five random seeds---123, 321, 12345, 42, 24---were used.  

\begin{table*}[h]
\caption{A list of hyperparameters. \textbf{m1-m9} are defined in the main text.}
\label{tab:app_hyperparam}
\centering\small
\begin{tabular}{c l c c c c}
        \toprule
        Hyperparameter & m1-m6 & m7 & m8 & m9  \\
        \midrule
        Node-wise latent variable dimension & 36 & 36 & 36 & 36 \\
        Atom neighbor cutoff [{\AA}] & 9.0 & 9.0 & 9.0 & 9.0 \\
        Residue neighbor cutoff [{\AA}] & 21.0 & 21.0 & 21.0 & 21.0 \\
        Encoder convolution depth & 3 & 3 & 3 & 3 \\
        Decoder convolution depth & 4 & 4 & 4 & 4 \\
        Maximum training hours [hr] & 20 & 20 & 20 & 20 \\
        Batch size & 4 & 4 & 4 & 4 \\
        Learning rate & 1e-3 & 1e-3 & 1e-3 & 1e-3 \\
        $\beta$ coefficient for KL divergence & 0.05 & 0.05 & 0.05 & 0.05 \\
        $\gamma$ coefficient for $L_{local}$ & 1.0 & 1.0 & 1.0 & 1.0 \\
        $\delta$ coefficient for $L_{torsion}$ & 1.0 & 0.0 & 1.0 & 1.0 \\
        $\eta$ coefficient for $L_{xyz}$ & 1.0 & 1.0 & 0.0 & 1.0 \\
        $\zeta$ coefficient for $L_{steric}$ & 3.0 & 3.0 & 3.0 & 0.0 \\
        \bottomrule
\end{tabular}
\end{table*}  

\subsection{Baseline (CGVAE)}
We modify the original version of CGVAE \cite{Wang2022Generative} to make it trainable for multiple chemical systems. Original CGVAE's encoder operates with atom-wise feature vectors, while GenZProt's encoder operates with residue-wise feature vectors. For a protein with $N$ residues and $n$ atoms, Original CGVAE's invariant encoder initializes $n$ node attributes with the atom identity. Then, it performs message passing operations through atom-atom pairs within a cutoff distance and pools the atom-wise information to obtain a CG bead-wise latent variable. Unlike GenZProt, the CGVAE encoder does not perform CG bead-CG bead pair message passing. CGVAE prior operates at the CG level---the prior initializes node feature vectors with the index of the corresponding CG bead and performs CG bead-CG bead pair message passing operations. When the model is trained for a single chemistry, the index alone would have provided enough information for all-atom reconstruction. However, for a transferable model, we provide additional information by initializing the node feature vector with the residue identity. For the encoder, we concatenate the residue identity with the atom identity to initialize the atom-wise feature vector. For the prior, we use residue identity to initialize the feature vector.

\section{Metrics}
\textbf{Root Mean Squared Distance (RMSD).} The reconstruction task evaluates the model's capacity to encode and reconstruct given structures. We report the $RMSD$ value of ground truth and reconstructed structures for each model. The lower the $RMSD$, the closer the generated structure is to the ground truth structure.   

\textbf{Graph Edit Distance (GED).} The sample quality is evaluated by measuring how well the generated geometries preserve the original chemical bond graph, which is quantified by the graph edit distance ratio $\lambda(G_{gen}, G_{true})$ between generated graph and the ground truth graph. $G_{gen}$ is deduced from the coordinates by connecting bonds between pair-wise atoms where the distances are within a threshold defined by an atomic covalent radius cutoff used in \cite{Wang2022Generative}. The lower the $\lambda$, the better $G_{gen}$ resembles $G_{true}$.   

\textbf{Steric clash score.} In addition to GED, we report the ratio of steric clash occurrence in all atom-atom pairs within a 5.0 {\AA} distance as a metric to measure the sample quality. For each atom-atom pair, distance smaller than 1.2 {\AA} is considered a steric clash.  

\section{Learning Objectives}

\subsection{Periodic angular loss}
\begin{figure}[h] 
\vskip 0.2in
\begin{center}
\centerline{\includegraphics[width=0.3\columnwidth]{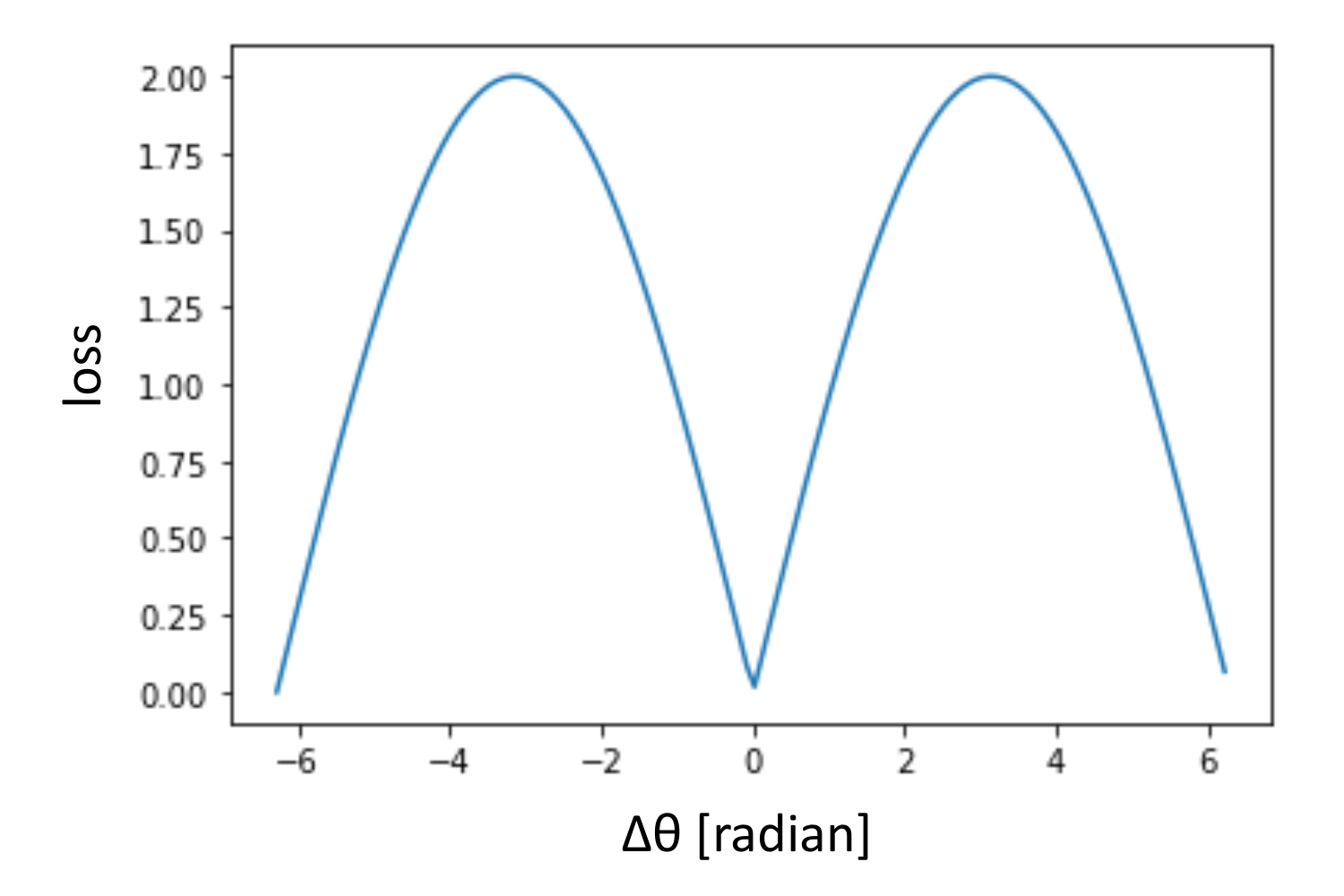}} 
\caption{Periodic angular loss. } 
\label{angle_loss}
\end{center}
\vskip -0.2in
\end{figure}

Periodic loss for angles introduced in Equation \ref{local_loss} is defined as:  
\begin{equation}
\begin{split}
  L_{\text{angle}} = \frac{1}{|A|}\sum_{\theta \in A} \sqrt{2 (1-\cos(\theta - \hat{\theta})) + \epsilon}
  \end{split} 
\end{equation}.  

This loss function is minimized at $\Delta\theta=\theta - \hat{\theta}=0, 2\pi$, and maximized at $\Delta\theta=\pi, 1.5\pi]$. Figure \ref{angle_loss} shows the angle loss term value as a function of $\Delta\theta$.  

\subsection{Interaction Score}
We devised an interaction score to evaluate the model's ability to learn long-range interactions. Interactions were identified based on atom-atom pairwise distances, as the distance is the most determining variable of intermolecular interactions: force field terms such as Lennard-Jones potential or electrostatic energy are computed as a function of distance. 

We tested adding the interaction score to our training objective, but the interaction score loss did not affect the model performance in reconstructing the long-range interactions. Thus, we introduce the score as a metric and not as a loss function.     

\textbf{Identification of the interactions. } We considered two classes of interactions.  
\begin{enumerate}
\item Hydrogen bonds, ion-ion interactions, dipole-dipole interactions : We identify heteroatom pairs within the distance of 3.3 {\AA}. 
\item Pi-pi stacking  : We identify a pair of aromatic rings (PHE, TYR, TRP, HIS) that the distance between their ring centers is smaller than 5.5 {\AA}.  
\end{enumerate}

The interaction score is defined as:  

\begin{equation}
  \begin{aligned}
  L_\text{atom-pair} \coloneqq \sum_{(\mathbf{x}, \mathbf{y}) \in \mathcal{A}} {\max(||\mathbf{x}-\mathbf{y}||^2_2 - 4.0,  0.0)} \\
  L_\text{pi-pair} \coloneqq \sum_{(\mathbf{x}, \mathbf{y}) \in \mathcal{P}} {\max(||\mathbf{x}-\mathbf{y}||^2_2 - 6.0,  0.0)} 
  \end{aligned}
\end{equation}

where $\mathcal{P}$ is a set of pair of atoms that are identified as type 1 interacting pair ($d_{xy} < 3.5$ {\AA}), and $\mathcal{P}$ is a set of pair of aromatic rings that are identified as type 2 interacting pair ($d_{xy} < 5.5$ {\AA}). The smaller the $L_\text{atom-pair}$ and $L_\text{pi-pair}$, the better the long-range interactions are reconstructed.  

Here, we report the interaction scores tested from different model architectures.  

\begin{table*}[h]
\caption{\textbf{Interaction scores}. \textbf{m1} : Our proposed model with equivariant encoder and invariant Z-matrix decoder. \textbf{m2} : Invariant encoder and Z-matrix decoder. \textbf{m3} : Equivariant encoder and Cartesian coordinate decoder. \textbf{m4} : Invariant encoder and Cartesian coordinate decoder. \textbf{m5} : \textbf{m1} trained with PED00151 only. \textbf{m6} : \textbf{m4} trained with PED00151 only. }
\label{tab:app_inter}
\centering\small
\begin{tabular}{c l cccc}
        \toprule
        & Method & PED00055 & PED00090 & PED00151 & PED00218   \\
        \midrule
        \rowcolor{Gray}
         \multirow{6}{*}{Interaction score ($\downarrow$)} 
         \cellcolor{white} & \textbf{m1 (GenZProt)}
         & \textbf{0.025}\scriptsize{$\pm$\textbf{0.000}}& \textbf{0.069}\scriptsize{$\pm$\textbf{0.002}} & \textbf{0.057}\scriptsize{$\pm$\textbf{0.000}} & \textbf{1.270}\scriptsize{$\pm$\textbf{0.000}}  \\
         & m2
         & 0.128\scriptsize{$\pm$0.003} & 0.282\scriptsize{$\pm$0.018} & 0.213\scriptsize{$\pm$0.003} & 1.332\scriptsize{$\pm$0.002}  \\
         & m3
         & 2.527\scriptsize{$\pm$0.165} & 1.539\scriptsize{$\pm$0.014} & 2.139\scriptsize{$\pm$0.085} & 2.412\scriptsize{$\pm$0.006}  \\
         & m4 (CGVAE) 
         & 1.416\scriptsize{$\pm$0.202} & 1.141\scriptsize{$\pm$0.043} & 1.797\scriptsize{$\pm$0.555} & 1.593\scriptsize{$\pm$0.215}  \\
         & m5 (GenZProt, single)
         & - & - & 0.221\scriptsize{$\pm$0.001} & - \\
         & m6 (CGVAE, single)
         & - & - & 1.574\scriptsize{$\pm$0.016} & - \\
        \bottomrule
\end{tabular}
\end{table*}

%%%%%%%%%%%%%%%%%%%%%%%%%%%%%%%%%%%%%%%%%%%%%%%%%%%%%%%%%%%%%%%%%%%%%%%%%%%%%%%
%%%%%%%%%%%%%%%%%%%%%%%%%%%%%%%%%%%%%%%%%%%%%%%%%%%%%%%%%%%%%%%%%%%%%%%%%%%%%%%

\end{document}